# Constrained Joint Cascade Regression Framework for Simultaneous Facial Action Unit Recognition and Facial Landmark Detection

Yue Wu    Qiang Ji
ECSE Department, Rensselaer Polytechnic Institute
110 8th street, Troy, NY, USA
{wuy9, jiq}@rpi.edu

## Abstract

*Cascade regression framework has been shown to be effective for facial landmark detection. It starts from an initial face shape and gradually predicts the face shape update from the local appearance features to generate the facial landmark locations in the next iteration until convergence. In this paper, we improve upon the cascade regression framework and propose the Constrained Joint Cascade Regression Framework (CJCRF) for simultaneous facial action unit recognition and facial landmark detection, which are two related face analysis tasks, but are seldomly exploited together. In particular, we first learn the relationships among facial action units and face shapes as a constraint. Then, in the proposed constrained joint cascade regression framework, with the help from the constraint, we iteratively update the facial landmark locations and the action unit activation probabilities until convergence. Experimental results demonstrate that the intertwined relationships of facial action units and face shapes boost the performances of both facial action unit recognition and facial landmark detection. The experimental results also demonstrate the effectiveness of the proposed method comparing to the state-of-the-art works.*

## 1. Introduction

Facial action unit recognition and facial landmark detection are two important tasks for face analysis. Facial action unit recognition refers to the automatic estimation of the Action Units (AU) defined in the Facial Action Coding System (FACS) [3]. FACS system and the facial action units provide objective measurements of the facial muscle movements and the facial motions. Facial landmark detection refers to the localization of the facial key points, such as the eye corners and the nose tip on facial images. The locations of the detected landmarks characterize the face shape. Both facial action unit recognition and facial landmark detection

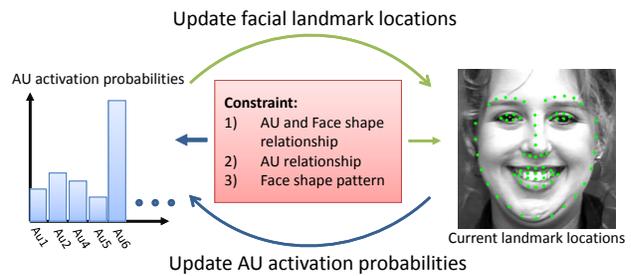

Figure 1. Constrained joint cascade regression framework for simultaneous facial action unit recognition and landmark detection.

would enable the machine understanding of human facial behavior, intent, emotion etc.

Facial action unit recognition and facial landmark detection are related tasks, but they are seldomly exploited jointly in the literatures. For example, the face shape defined by the landmark locations are considered as effective features for AU recognition. But, the landmark location information is usually extracted beforehand with facial landmark detection algorithms. On the other hand, the Action Unit information is rarely utilized in the literature to help facial landmark detection, even though the facial muscle movements and the activation of specific facial action unit can cause the appearance and shape changes of the face which significantly affect facial landmark detection. The mutual information and intertwined relationship among facial action unit recognition and facial landmark detection should be utilized to boost the performances of both tasks.

Cascade regression framework has been shown to be an effective method for face alignment recently [19][13]. It starts from an initial face shape (e.g. mean face) and iteratively updates the facial landmark locations based on the local appearance features until convergence. Several regression models have been applied to learn the mapping from the local appearance features to the face shape update.

To leverage the success of the cascade regression framework and to achieve the goal of joint facial action unit recognition and landmark detection, we proposed the Constrained Joint Cascade Regression Framework (CJCRF). The general framework of the proposed method is shown in Figure 1. First, we learn the relationship among facial shapes and facial action units as a constraint before the cascade training. Second, in the constrained joint cascade regression framework, with the help from the constraint, we iteratively update the landmark locations and the action unit activation probabilities. When updating the landmark locations, we use the local appearance information, the currently estimated AU activation probabilities, and the pre-trained constraint that captures the joint relationship among AUs and face shapes. When updating the AU activation probabilities, we use the local appearance information, currently estimated landmark locations, and the pre-trained constraint that captures the relationship among AUs and face shapes. The landmark detection and AU estimation interact to reach convergence.

The major contributions of the proposed method are as follows:

- **Joint cascade regression framework:** We improve upon the effective cascade regression framework and propose the Constrained Joint Cascade Regression Framework (CJCRF). CJCRF jointly performs facial landmark detection and facial action unit recognition, which is novel comparing to most of the existing methods that estimate them separately.
- **Constraint:** CJCRF explicitly utilizes the relationship among facial action units and face shapes as a constraint to improve both the facial action unit recognition and facial landmark detection. In addition, the AU relationship and face shape patterns are embedded in the constraint.
- **Experiments:** The experiments show that the joint cascade regression framework boosts the performances of both facial action unit recognition and facial landmark detection, comparing to the state-of-the-art works.

The remaining part of the paper is organized as follows. In section 2, we review the related work. In section 3, we discuss the proposed method. In section 4, we evaluate the proposed method and compare it to state-of-the-art works. We conclude the paper in section 5.

## 2. Related Work

### 2.1. Facial landmark detection

The facial landmark detection algorithms can be classified into the holistic methods [1][11], the Constrained Local Method (CLM) [2], and the regression based methods [15][19]. The holistic methods learn holistic appearance and face shape models, while the CLM learns the holistic shape model with local appearance model. The regression based methods directly learn the mapping from image appearance to the face shape without explicit appearance and shape models.

The regression based methods can be classified into the direct mapping methods and the cascade regression framework. The direct mapping methods directly map the image appearance to the absolute landmark locations [15]. The cascade regression framework starts from an initial face shape (e.g. mean face shape), and it gradually updates the landmark locations from the local appearance information collected around the currently predicted landmark locations to generate the facial landmark locations in the next iteration until convergence. The cascade regression based methods usually differ in the features and the regression models. For example, in [19], SIFT features and linear regression model is used. In [13], local binary features learned from the regression tree models are combined with the linear regression models to predict the landmark locations. The proposed method follows the cascade regression framework.

### 2.2. Facial action unit recognition

The facial action unit recognition algorithms usually focus on the improvements of the features or the classifier design. The features can be classified into appearance features and shape features. The popular appearance features include the LBP features [7], the Local Gabor Binary Pattern (LGBP) features [14], and the Discrete Cosine Transform (DCT) features [5]. The shape features are extracted from the facial landmark locations, and typical shape features are the absolute landmark locations, the distance between pairs of points, and the angles defined by a set of three points. In terms of classifier design, the facial action units can be recognized independently or jointly with machine learning techniques. Comparing to the independent AU recognition methods [10] that ignore the AU relationship, the joint methods usually can achieve better performance by adding the AU relationship or dynamic dependencies. For example, in [4], the AU relationship is embedded in the Multi-conditional Latent Variable Model (MC-LVM). In [18], the global AU relationship is modeled by the Hierarchical Restricted Boltzmann Machine (HRBM) model. In [20], joint AU recognition task is formulated as a Multi-Task Multiple Kernel Learning (MTMKL) problem. In [22], pairwise AU relationships are learned. In [8], the AU relationship is learned from prior knowledge. In [17], the temporal information is incorporated for AU recognition.

Although facial action unit recognition and facial landmark detection are related tasks, their interaction is usually one way in the aforementioned methods, i.e. facial landmark locations are extracted as features for facial action unit recognition. [9] is the most similar work that jointly performs facial landmark tracking and facial action unit recog-

**Algorithm 1**: General Cascade Regression Framework

Initialize the landmark locations $\mathbf{x}^0$ using the mean face.
**for** *t=1, 2, ..., T or convergence* **do**
  Update the landmark locations, given the image and the current landmark location.
  $$f_t : \mathbf{I}, \mathbf{x}^{t-1} \to \Delta \mathbf{x}^t$$
  $$\mathbf{x}^t = \mathbf{x}^{t-1} + \Delta \mathbf{x}^t$$
**end**
Output the estimated landmark locations $\mathbf{x}^T$.

---

**Algorithm 2**: Constrained Joint Cascade Regression Framework (CJCRF)

**Learn the constraint:**
Learn the joint relationship among facial landmark locations and action unit labels, the AU relationship, face shape patterns as a constraint, denoted as $\mathbb{C}(.)$.
**Constrained Joint Cascade Regression Framework:**
Initialize the landmark locations $\mathbf{x}^0$ using the mean face;
Initialize the AU activation probability as 0.5, $p_i^0 = 0.5, \forall i$
**for** *t=1, 2, ..., T or convergence* **do**
  Update the landmark locations, given the image, the current landmark location, the current AU activation probabilities, and the constraint.
  $$f_t : \mathbf{I}, \mathbf{x}^{t-1}, \mathbf{p}^{t-1}, \mathbb{C}(.) \to \Delta \mathbf{x}^t$$
  $$\mathbf{x}^t = \mathbf{x}^{t-1} + \Delta \mathbf{x}^t$$
  Update the AU activation probabilities, given the image, the current landmark location, and the constraint.
  $$g_t : \mathbf{I}, \mathbf{x}^t, \mathbb{C}(.) \to \Delta \mathbf{p}^t$$
  $$\mathbf{p}^t = \mathbf{p}^{t-1} + \Delta \mathbf{p}^t$$
**end**
Output the estimated landmark locations $\mathbf{x}^T$ and the AU activation probabilities $\mathbf{p}^T$.

---

nition by building a hierarchical Dynamic Bayesian Network to capture their joint relationship. Our model differs from [9]. While they capture local dependencies, our model captures global AU relationship, global face shape patterns, and global dependencies among AU and facial landmarks. While they perform one-time prediction, we iteratively update the AU activation probabilities and landmark locations to achieve robustness. In [21], facial expression recognition is considered as an auxiliary task to help the learning of better feature representations for facial landmark detection. The auxiliary tasks do not include action unit recognition, and their performances are not reported. In contrast, we aim to exploit the interdependencies between AU and landmarks to improve both tasks. In addition, we require less training data, comparing to this deep method.

## 3. Constrained Joint Cascade Regression Framework

In this section, we first review the general cascade regression framework and then introduce the proposed Constrained Joint Cascade Regression Framework (CJCRF).

### 3.1. General cascade regression framework

Before we discuss the proposed method, we first review the general cascade regression framework, which has been successfully applied to facial landmark detection [19][13]. The overall algorithm is shown in Algorithm 1. Assume that the facial landmark locations are denoted as $\mathbf{x} \in \Re^{2D}$, where $D$ is the number of facial landmark points. The test image is denoted as $\mathbf{I}$. Starting from the mean face $\mathbf{x}^0$, the cascade regression method iteratively predicts the face shape update $\Delta \mathbf{x}^t$ based on the local appearance features with regression model, and adds the face shape update to the current shape $\mathbf{x}^{t-1}$ to generate the shape $\mathbf{x}^t$ for the next iteration until convergence.

The existing methods use different image features and regression functions in the cascade regression framework. For example, in the Supervised Descent Method (SDM) [19], linear regression model is used to learn the prediction.

$$f_t : \Delta \mathbf{x}^t = \mathbf{R}^t \Phi(\mathbf{x}^{t-1}, \mathbf{I}), \quad (1)$$

where $\Phi(\mathbf{x}^{t-1}, \mathbf{I}) \in \Re^{128D}$ denotes the local appearance features (e.g. SIFT) centered at the currently predicted landmark locations, and $\mathbf{R}^t$ is the parameter of the linear regression model.

For training, given the training data $\{\mathbf{x}_m^*, \mathbf{I}_m\}$, where $\mathbf{x}_m^*$ represents the ground truth facial landmark locations, the ground truth face shape update can be calculated as $\Delta \mathbf{x}_m^{t,*} = \mathbf{x}_m^* - \mathbf{x}_m^{t-1}$. Then, the model parameter $\mathbf{R}^t$ can be estimated in a least square formulation and solved with the closed form solution:

$$\mathbf{R}^t = arg \min_{\mathbf{R}} \sum_m \|\Delta \mathbf{x}_m^{t,*} - \mathbf{R}\Phi(\mathbf{x}_m^{t-1}, \mathbf{I}_m)\|_2^2. \quad (2)$$

### 3.2. Constrained joint cascade regression framework

The proposed Constrained Joint Cascade Regression Framework (CJCRF) improves upon the general cascade regression framework and it jointly performs facial landmark detection and facial action unit recognition. Assume the binary facial action unit labels are denoted as $\mathbf{a} \in \{0, 1\}^N$, where $N$ is the number of estimated facial action units. $p_i = P(a_i = 1; \mathbf{I})$ denotes the AU activation probability of action unit $a_i$ for the testing image $\mathbf{I}$. $\mathbf{p} = \{p_1, p_2, ..., p_N\}$ refers to the AU activation probability vector for all the AUs. The goal of CJCRF is to jointly estimate the landmark locations **x** and the AU activation probabilities **p** given the testing image, from which we estimate the AU labels **a**.

The general framework of CJCRF is shown in Figure 1 and Algorithm 2. First, we learn the joint relationship among facial landmark locations **x** and facial action unit labels **a**. Note that AU relationship and face shape patterns are also embedded. Second, in the constrained joint cascade regression framework, we iteratively update the landmark locations and the AU activation probabilities, with the help from the pre-learned joint relationship as a constraint (denoted as $\mathbb{C}(.)$). In particular, we initialize the landmark locations $\mathbf{x}^0$ using the mean face, and initialize the initial AU activation probabilities $p_i^0 = 0.5, \forall i$. When updating the landmark locations, we predict the face shape update $\Delta \mathbf{x}^t$ based on the currently estimated landmark locations $\mathbf{x}^{t-1}$, the AU activation probabilities $\mathbf{p}^{t-1}$, and the constraint $\mathbb{C}(.)$. The estimated shape update is added to the current face shape to generate the landmark locations for the next iteration. When updating the AU activation probabilities, we predict the probability update $\Delta \mathbf{p}^t$ based on the currently estimated landmark locations $\mathbf{x}^t$ and the constraint $\mathbb{C}(.)$. The probability update is then added to the currently estimated AU activation probabilities.

One important property of the proposed method is that we iteratively update both the landmark locations and AU activation probabilities. The iterative procedure would accumulate information from all the iterations to achieve robustness comparing to one-time estimation. In the following, we discuss the three major components of the proposed method, including the joint relationship constraint, landmark location prediction, and AU activation probability prediction.

### 3.2.1 AU and facial landmark relationship constraint

We learn the joint relationship among face shapes and the binary action unit labels using the Restricted Boltzmann Machine model (RBM). As shown in Figure 2, RBM model captures the joint probability of visible variables through several binary hidden nodes **h**. In our application, the visible nodes are the continues facial landmark locations **x** and the binary facial action unit labels **a**:

$$P(\mathbf{a}, \mathbf{x}; \theta) = \frac{\sum_{\mathbf{h}} e^{-E(\mathbf{a}, \mathbf{x}, \mathbf{h}; \theta)}}{Z(\theta)} \quad (3)$$

$$E(\mathbf{a}, \mathbf{x}, \mathbf{h}; \theta) = \sum_j \frac{(x_j - b_{\mathbf{x},j})^2}{2} - \mathbf{b}_\mathbf{a}^T \mathbf{a} - \mathbf{c}^T \mathbf{h} \\ - \mathbf{x}^T \mathbf{W}_\mathbf{x} \mathbf{h} - \mathbf{a}^T \mathbf{W}_\mathbf{a} \mathbf{h} \quad (4)$$

$$Z(\theta) = \sum_{\mathbf{a}, \mathbf{x}, \mathbf{h}} e^{-E(\mathbf{a}, \mathbf{x}, \mathbf{h}; \theta)} \quad (5)$$

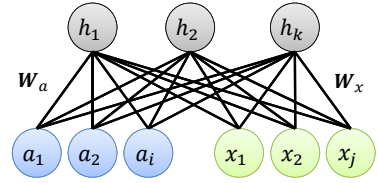

Figure 2. Restricted Boltzmann Machine (RBM) model that embeds the joint relationships among AUs and landmark, the AU relationship, and the face shape patterns.

Here, $E(\mathbf{a}, \mathbf{x}, \mathbf{h}; \theta)$ is the energy function, and $Z(\theta)$ is the partition function. The parameter $\theta = \{\mathbf{W_x}, \mathbf{W_a}, \mathbf{b_x}, \mathbf{b_a}, \mathbf{c}\}$ includes the pairwise parameters $\mathbf{W_x} \in \Re^{2D \times K}$, $\mathbf{W_a} \in \Re^{N \times K}$, and the biases $\mathbf{b_x} \in \Re^{2D}$, $\mathbf{b_a} \in \Re^N$, $\mathbf{c} \in \Re^K$. $K$ is the number of hidden variables. Given the ground truth pairwise training data with AU labels and landmark locations $\{\mathbf{a}_m, \mathbf{x}_m\}_{m=1}^M$, model training can be performed with Contrastive Divergence algorithm (CD) [6].

The model captures three levels of relationships. First, the joint probability $P(\mathbf{a}, \mathbf{x}; \theta)$ captures the global and joint relationship among facial landmark locations and facial action unit labels. Second, the AU relationship is embedded in the marginal distribution $P(\mathbf{a}; \theta)$. Third, the face shape patterns are also embedded in the marginal distribution $P(\mathbf{x}; \theta)$. Those three levels of information can be used in two ways in the constrained joint cascade regression framework. On one hand, once we know the activation of particular action unit (e.g. $a_i = 1$), we could find out the face shape that is consistent with the AU, and this information should be included to help the facial landmark detection. In particular, we use the AU dependent expected face shape, where the the expectation is taken over the marginal conditional probabilities $P(\mathbf{x}|a_i = 1; \theta)$ embedded in the RBM model.

$$\mathbb{E}_{P(\mathbf{x}|a_i=1;\theta)}[\mathbf{x}] = \int_\mathbf{x} \mathbf{x} P(\mathbf{x}|a_i = 1; \theta) d\mathbf{x} \quad (6)$$

The AU dependent shapes characterize the common properties of the face shapes that have the activation of particular AU, and they change with different AUs. For example, as shown in Figure 3, given the activation of different AUs, we have different prior knowledge of the face shapes. On the other hand, once we have some knowledge of the current face shape **x**, we could also estimate the AU activation probabilities from the model $P(a_i = 1|\mathbf{x}; \theta)$, and this information should be incorporated for AU recognition. For example, Figure 4 shows the estimated AU activation probabilities given one particular face shape from the model. It can be seen that, the face shape would provide distinct knowledge about the activations of AUs. In the following, we further explain how to use the learned joint relationship as a constraint to help the predictions of action unit activation probability and landmark locations.

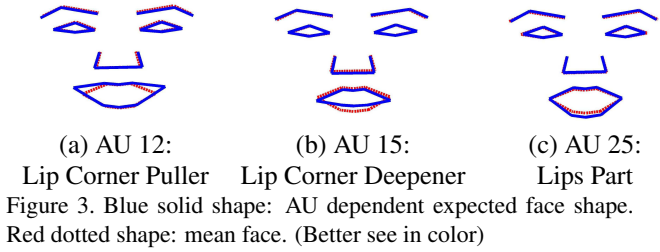

(a) AU 12: Lip Corner Puller  (b) AU 15: Lip Corner Deepener  (c) AU 25: Lips Part

Figure 3. Blue solid shape: AU dependent expected face shape. Red dotted shape: mean face. (Better see in color)

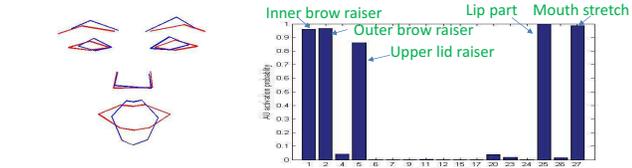

Figure 4. Estimated AU activation probabilities (right) given the facial shape (left). Blue: current face shape. Red: mean face.

### 3.2.2 Update the landmark locations with constraint

Given the learned relationships among facial landmark locations and facial action unit labels as a constraint, the constrained cascade regression method iteratively predicts the facial landmark locations and the facial action unit activation probabilities. When updating the landmark locations, the method predicts the facial shape update based on the currently estimated landmark locations $\mathbf{x}^{t-1}$, the AU activation probabilities $\mathbf{p}^{t-1}$, and the RBM model with parameter $\theta$ that embeds the relationships. It is formulated as a general optimization problem:

$$\begin{aligned}\underset{\Delta \mathbf{x}^t}{\text{minimize}} \quad & \|\Delta \mathbf{x}^t - \mathbf{R}^t \Phi(\mathbf{x}^{t-1}, \mathbf{I})\|_2^2, \\ \text{subject to} \quad & \mathbf{x}^t = \bar{\mathbf{x}}(\mathbf{p}^{t-1}, \theta)\end{aligned} \quad (7)$$

where $\mathbf{x}^t = \mathbf{x}^{t-1} + \Delta \mathbf{x}^t$. The formulation leverages two information resources. In the objective function, similar to the conventional cascade regression method in Equation 1, we predict the face shape update $\Delta \mathbf{x}^t$ from the local appearance features $\Phi(\mathbf{x}^t - 1, \mathbf{I})$ with linear regression model. In addition, we explicitly add the constraint to ensure that the predicted face shape is consistent with the currently predicted AU activation probabilities $\mathbf{p}^{t-1}$. In particular, $\bar{\mathbf{x}}(\mathbf{p}^{t-1}, \theta)$ denotes the estimation of the facial landmark locations from the the joint relationship embedded in RBM based on the currently estimated AU activation probabilities.

$$\bar{\mathbf{x}}(\mathbf{p}^{t-1}, \theta) = \sum_i \mathbb{E}_{P(\mathbf{x}|a_i=1;\theta)}[\mathbf{x}] \frac{p_i^{t-1}}{\sum_l p_l^{t-1}} \quad (8)$$

Here, $\mathbb{E}_{P(\mathbf{x}|a_i=1;\theta)}[\mathbf{x}]$ is the AU dependent expected face shape, calculated from the RBM model that captures the relationship among AUs and facial shape in Equation 6. To take into account of the currently estimated AU activation probabilities, those AU dependent expected face shape are combined with weights that are based on the currently estimated AU activation probabilities. The intuition is to have higher weight for specific AU dependent expected face shape if the activation probability of the particular AU is high in the current testing image.

In inference, with lagrangian relaxation, Equation 7 becomes a standard least square problem, and it can be solved in the closed form solution. Model training refers to the learning of the linear regression parameter $\mathbf{R}^t$. Given the training data, the learning of $\mathbf{R}^t$ is similar to that in the general cascade regression framework as shown in Equation 2.

### 3.2.3 Update the action unit activation probabilities with constraint

When updating the AU activation probabilities, we predict the AU activation probability update from the current landmark locations $\mathbf{x}^t$, the current AU activation probabilities $\mathbf{p}^{t-1}$, and the RBM model with parameter $\theta$. We formulate this as a general optimization problem:

$$\begin{aligned}\underset{\Delta \mathbf{p}^t}{\text{minimize}} \quad & \|\Delta \mathbf{p}^t - \mathbf{T}^t \Phi(\mathbf{x}^t, \mathbf{I})\|_2^2, \\ \text{subject to} \quad & p_i^t = P(a_i = 1|\mathbf{x}^t; \theta), \forall i \\ & \mathbf{0} \le \mathbf{p}^t \le \mathbf{1}\end{aligned} \quad (9)$$

where $\mathbf{p}^t = \mathbf{p}^{t-1} + \Delta \mathbf{p}^t$. The prediction of AU activation probabilities utilizes two information resources. In the objective function, we use linear regression model with parameter $\mathbf{T}^t$ to predict the AU activation probability update from the local appearance features $\Phi(\mathbf{x}^t, \mathbf{I})$. In the constraint, we ensure that the predicted AU activation probability $p_i^t$ is consistent with the current landmark locations $\mathbf{x}^t$. In particular, given the current shape $\mathbf{x}^t$, we could estimate the AU activation probabilities from the RBM model that captures the relationship among AUs and facial shape using $P(a_i = 1|\mathbf{x}^t; \theta)$ as illustrated in section 3.2.1. The estimated AU activation probability of the same AU $a_i$ for the current testing image, denoted as $p_i^t$, should be close to the prediction from the prior RBM model.

In inference, with the lagrangian relaxation, the optimization problem in Equation 9 becomes a least square problem and it can be solved with the closed form solution. Model training refers to the learning of the linear regression function $\mathbf{T}^t$, and the estimation is similar to that shown in Equation 2. In particular, given the training data, the ground truth probabilities $\mathbf{p}_m^*$ can be generated based on the ground truth AU labels $\mathbf{a}_m^*$ for image $\mathbf{I}_m$ ($p_i^* = P(a_i = 1) = 1$ if $a_i^* = 1$). The ground truth probability update can be calculated as $\Delta \mathbf{p}_m^{t,*} = \mathbf{p}_m^* - \mathbf{p}_m^{t-1}$. Then the parameter estimation is formulated as a least square problem with the closed form solution:

$$\mathbf{T}^t = arg\min_{\mathbf{T}} \sum_m \|\Delta \mathbf{p}_m^{t,*} - \mathbf{T}\Phi(\mathbf{x}_m^t, \mathbf{I}_m)\|_2^2 \quad (10)$$

## 4. Experimental Results

### 4.1. Implementation details

**Database:** We evaluate the proposed method on both the posed and spontaneous data. We use the Extended Cohn-Kanade database(CK+) [10] as the posed data. It contains 593 facial activity videos from 210 subjects with various ethnicities. The peak frames are FACS coded with 30 facial action units. We manually labeled 28 facial landmarks on the peak frames from each sequence and performed the evaluations on the peak frames. To evaluate the experiments on spontaneous data, we use the SEMAINE database [12] and the FERA database [16], in which the facial sequences undergo significant variations of speech related movements, head poses, etc. In the SEMAINE database, the emotionally colored conversation between the human subjects and machine agents is recorded, where the emotion related facial activity of subjects are naturally induced by the agent. FACS are coded for 180 images from 8 sessions. We also manually labeled 28 facial landmarks on those frames. The FERA database contains 87 AU related sequences of 7 actors, and conversion is also involved. 50 AU annotations for each frame are provided and we manually annotated 28 facial landmarks on 260 frames for the experiments.

**Model:** The RBM model that learns the joint relationship among AUs and landmarks, contains 150 hidden node, and we train it with 800 epochs. To calculate the local appearance features, we use SIFT and set the local region with a radius about 0.166 of the face size. There are four iterations in the cascade framework. Following the previous cascade regression method [19], we augment the training images by adding random scale, rotation, and translation perturbations to the initial face shape. When calculating the AU dependent face shapes in Equation 6, we average the training data with activated corresponding AU. The lagrangian relaxation parameters are set to be 0.5.

**Evaluation criteria:** To calculate the facial landmark detection error, we use the distance between the detected point and the ground truth point normalized by the interocular distance. We calculate the mean error by averaging over all the available points and testing images. To evaluate the facial action unit recognition accuracy, we use the $F_1$ score and the area under the ROC curve (AUC). Similar to the previous work [10], the score is weighted by the frequency of the presents of AUs.

### 4.2. Performance for posed facial actions

#### 4.2.1 Evaluation of the proposed method

Depending on whether we use the joint relationships among facial shape and AUs as equality constraints in Equation 7 and 9, there are four versions of the proposed method. In (1) **CJCRF-NoConstraint**, there is no equal-ity constraint. Facial landmark detection is performed independently without AU information. AU recognition is only based on the local appearance information. In (2) **CJCRF-ConstraintLandmark**, AU information is incorporated through the constraint to help landmark detection. In (3) **CJCRF-ConstraintAU**, facial shape information is incorporated through the constraint to help AU recognition, comparing to (1). In (4) **CJCRF**, the joint relationships among landmark locations and AUs are incorporated to help both tasks. Here, we evaluate the four versions of the proposed method on the posed CK+ database with five-folder cross validation and mutually excluded subjects.

Table 1. Comparison of different versions of the proposed methods. See text for the descriptions of each version.

|  | detection error | AU recognition $F_1$(%) | AUC |
|---|---|---|---|
| CJCRF-NoConstraint | 5.13 | 70.80 | 89.86 |
| CJCRF-ConstraintLandmark | 4.86 | 72.02 | 90.20 |
| CJCRF-ConstraintAU | 5.18 | 78.72 | 94.17 |
| CJCRF | 4.86 | 78.74 | 94.28 |

The experimental results are shown in Table 1. There are a few observations. First, without the constraint on AU and Landmark detection, the baseline method (**CJCRF-NoConstraint**) performs the worst. As shown in Figure 5, without the AU information, the standard cascade regression landmark detection algorithm [19] would fail with limited training data. Second, comparing **CJCRF-ConstraintLandmark** to **CJCRF-NoConstraint**, by adding the constraint to landmark prediction, the landmark detection error is significantly reduced with the slight improvement of AU recognition accuracy. Third, comparing **CJCRF-ConstraintAU** to **CJCRF-NoConstraint**, by adding the constraint to AU prediction, the AU recognition performance improves obviously. Lastly, with the constraint on both the landmark and AU prediction, the full model (**CJCRF**) achieves the best performances.

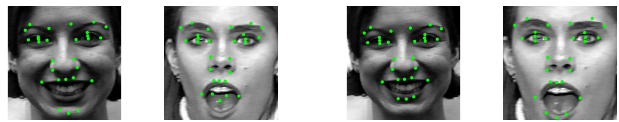

Figure 5. Left two images: landmark detections without AU info. Right two images: proposed method with AU info.

Figure 6 shows that by using the proposed full **CJCRF** method, both the facial landmark detection and AU recognition performances increase over cascade iterations and they converge quickly. With Matlab implementation, the full model achieves 5 fps on a single core machine.

#### 4.2.2 Comparison with state-of-the-arts

To compare the proposed method to the state-of-the-art works, we retrain the full model (**CJCRF**) with leave-one-

3405

Table 2. Comparison of the proposed method to state-of-the-art works on CK+ database [10]. "*" denotes the reported results from the original paper. "(.)" denotes the results with different experimental settings.

| | detection error | 17 AU $F_1(\%)$ | 17 AU AUC | 13 AU $F_1(\%)$ | 13 AU AUC | 8 AU $F_1(\%)$ | 8 AU AUC |
|---|---|---|---|---|---|---|---|
| SDM [19] | 4.96 | - | - | - | - | - | - |
| Lucey (AAM, SVM)[10] | 6.12 | - | 94.5* | - | - | - | - |
| HRBM[18] | - | 79.21* | - | - | - | - | - |
| DBN[9] | (2.59)* | - | - | 77.26* | - | - | - |
| Temporal[17] | - | - | - | 61.59 *(ck) | - | - | - |
| MC-LVM [4] | - | - | - | - | - | 80.14* | - |
| $l_p$-MTMKL [20] | - | - | - | - | - | 73.93* | - |
| CJCRF (proposed) | **4.66** | **80.72** | **94.92** | **83.41** | **95.67** | **80.76** | **94.37** |

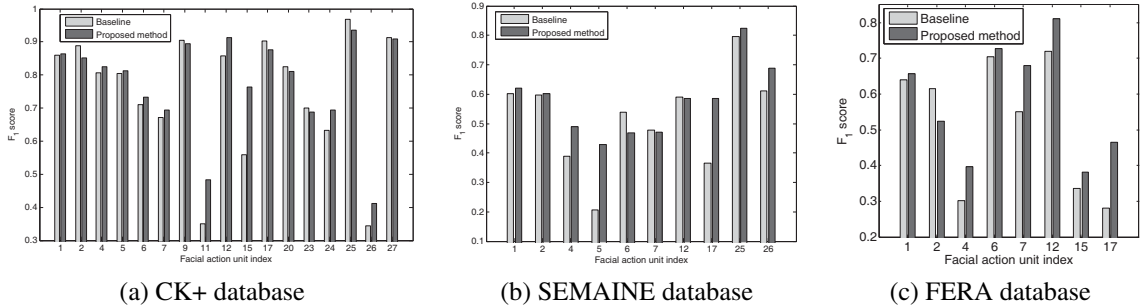

(a) CK+ database    (b) SEMAINE database    (c) FERA database

Figure 7. $F_1$ scores for individual action unit on different databases. The baseline methods are: (a) SVM [10], (b) SVM, (c) Data-free [8].

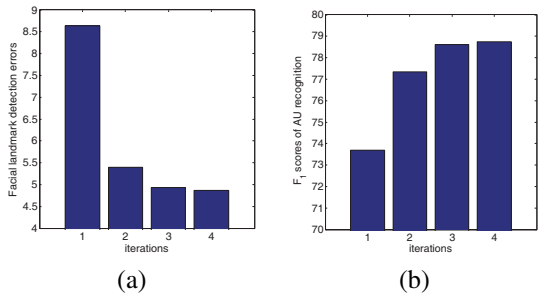

(a)    (b)

Figure 6. Performances of the proposed method over cascade iterations. (a) Landmark detection errors. (b) AU recognition, $F_1$ scores.

subject-out cross validation setting, following the previous works [18]. Since different algorithms may be evaluated on different AUs, we show the performances of the proposed method on three AU sets (trained also with those AUs). The **17 AU** set contains AU 1, 2, 4, 5, 6, 7, 9, 11, 12, 15, 17, 20, 23, 24, 25, 26 and 27. The **13 AU** set contains AU1, 2, 4, 5, 6, 7, 9, 12, 15, 17, 24, 25, and 27. The **8 AU** set contains AU1, 2, 4, 6, 7, 12, 15 and 17. For each AU set, the baseline methods may either predict only the AUs in the set, or predict more AUs but share the common AUs as the proposed method.

The experimental comparison results are shown in Table 2. Although the performances on CK+ become saturate, especially for AU recognition, our algorithm still can improve the accuracies. For facial landmark detection, the proposed method significantly outperforms the Active Ap-pearance Model as used in [10], and the Supervised Descent Method (SDM) [19]. The results from the DBN [9] are not directly comparable, since their model was evaluated on a particular data subset, which is not accessible. Figure 8 (a) shows more sample images with landmark detection results on the CK+ database. For AU recognition, there are a few observations. First, on **17 AU** set, the performance of the proposed method is better than [10] that uses SVM, and the HRBM [18]. We also implemented another baseline AU recognition method that uses SVM and concatenate SIFT features extracted around detected landmark locations with SDM algorithm for each AU independently. Its $F1$ score and $AUC$ are 71.33% and 91.03% respectively. Second, on the **13 AU** set and **8 AU** set, the proposed method achieves the best performance. The recognition performance on each individual AU using the proposed method and the baseline method [10] is shown in Figure 7 (a).

### 4.3. Performance for non-posed facial actions

In this section, we evaluate the proposed constrained joint cascade regression method (full model **CJCRF**) on the spontaneous SEMAINE database [12] and FERA database [16]. For the SEMAINE database, similar as the previous work [18], we perform facial action unit recognition of the 10 most frequent AUs: 1, 2, 4, 5, 6, 7, 12, 17, 25, and 26. For the FERA database, we evaluate the 8 most frequent AUs: 1, 2, 4, 6, 7, 12, 15, and 17.

The experimental results on SEMAINE database are shown in Table 3. We evaluate the proposed method in two settings. In the first cross database setting, we train the model on CK+ database, and test it on the SEMAINE database. In the second setting, we perform leave-one-subject-out cross validation on the SEMAINE database. There are a few observations. First, with the cross database setting, for facial landmark detection, CJCRF is significantly better than SDM [19] as an conventional cascade regression method. For AU recognition, the proposed CJCRF method outperforms the baseline SVM algorithm and is comparable to the HRBM method [18]. One possible reason is that HRBM uses manually labeled eye locations to extract the features, while we rely on automatically detected facial landmarks. Second, with the leave-one-subject-out cross validation setting, the proposed method is better than SVM as the baseline. Third, the performance of all methods drop significantly on the difficult SEMAINE database.

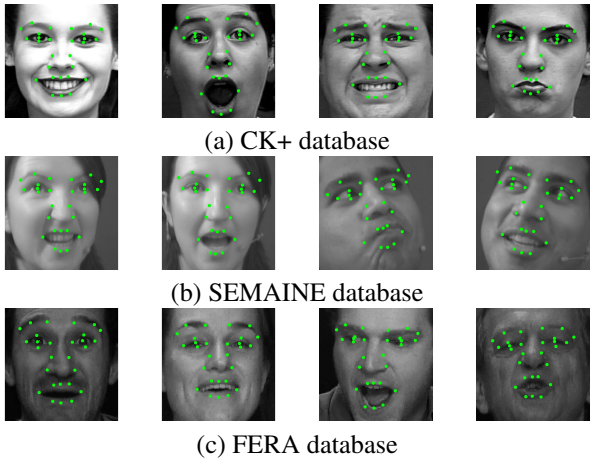

(a) CK+ database

(b) SEMAINE database

(c) FERA database

Figure 8. Facial landmark detection results using the proposed method on sample images from different databases.

Table 3. Comparison of the proposed method to state-of-the-arts on SEMAINE database [12]. Cross database: train on CK+, test on SEMAINE. Within database: cross subject validation on SEMAINE. "*" denotes the reported results from the original paper.

|  | Cross database | | Within database | |
| --- | --- | --- | --- | --- |
|  | detection | AU ($F_1$) | detection | AU ($F_1$) |
| SDM [19] | 11.06 | - | 12.46 | - |
| SVM | - | 47.70 | - | 51.74 |
| HRBM [18] | - | **54.76*** | - | - |
| CJCRF (proposed) | 8.28 | 52.22 | 10.38 | **56.68** |

The experimental results on FERA database are shown in Table 4. With the same leave-one-subject-out cross validation setting as the previous work [8], the proposed method outperforms the LGBP [14], the Data-free Model [8], the HRBM [18] method, and DCT [5] for facial action unit recognition. For facial landmark detection, the proposed method outperforms the SDM [19].

In Figure 7 (b)(c), we plot the $F_1$ scores of AU recognition performance on each individual AU for the spontaneous

Table 4. Comparison of the proposed method to state-of-the-art works on FERA database [16]. "*" denotes the reported results from the original paper.

|  | detection error | AU recognition $F_1$(%) |
| --- | --- | --- |
| SDM [19] | 6.17 | - |
| LGBP [14] | - | 46.24* |
| Data-free [8] | - | 52.62* |
| HRBM [18] | - | 54.60* |
| DCT [5] | - | 56.58* |
| CJCRF (proposed) | **5.90** | **59.66** |

databases. Figure 8 (b)(c) show some sample images with landmark detection results.

## 5. Conclusion

In this paper, we proposed the Constrained Joint Cascade Regression Framework (CJCRF) that improves over the existing effective cascade regression method. The CJCRF method jointly performs facial action unit recognition and facial landmark detection. The model first learns the joint relationship among facial action units and face shapes as a constraint. Then, in the constrained joint cascade regression framework, with the help from the constraint, the method iteratively updates the facial landmark locations and the AU activation probabilities until convergence. The experiments demonstrate that the intertwined relationship of facial action units and face shapes boost the performances of both facial landmark detection and facial action unit recognition. They also demonstrate the effectiveness of the proposed method comparing to state-of-the-art works on both poses and spontaneous databases.

In the future, we would extend the proposed method to perform AU recognition on images with large head movements. One possible direction is to decouple the rigid head movement and non-rigid deformation before estimation. Furthermore, the proposed method can be improved using more advanced features and dynamic information.

**Acknowledgements:** The work described in this paper was supported in part by the Federal Highway Administration via grant DTFH6114C00005 and in part by the IBM Ph.D. Fellowship.

## References

[1] T. F. Cootes, G. J. Edwards, and C. J. Taylor. Active appearance models. *IEEE Transactions on Pattern Analysis and Machine Intelligence*, 23(6):681–685, June 2001. 2

[2] D. Cristinacce and T. Cootes. Automatic feature localisation with constrained local models. *Pattern Recognition*, 41(10):3054 – 3067, 2008. 2

[3] P. Ekman and W. Friesen. *Facial Action Coding System: A Technique for the Measurement of Facial Movement.* Consulting Psychologists Press, Palo Alto, 1978. 1

[4] S. Eleftheriadis, O. Rudovic, and M. Pantic. Multi-conditional latent variable model for joint facial action unit detection. In *Computer Vision (ICCV), 2015 IEEE International Conference on*, Dec 2015. 2, 7

[5] T. Gehrig and H. Ekenel. A common framework for real-time emotion recognition and facial action unit detection. In *Computer Vision and Pattern Recognition Workshops (CVPRW), 2011 IEEE Computer Society Conference on*, pages 1–6, June 2011. 2, 8

[6] G. E. Hinton. Training products of experts by minimizing contrastive divergence. *Neural Comput.*, 14(8):1771–1800, Aug. 2002. 4

[7] B. Jiang, M. Valstar, and M. Pantic. Action unit detection using sparse appearance descriptors in space-time video volumes. In *Automatic Face Gesture Recognition and Workshops (FG 2011), 2011 IEEE International Conference on*, pages 314–321, March 2011. 2

[8] Y. Li, J. Chen, Y. Zhao, and Q. Ji. Data-free prior model for facial action unit recognition. *IEEE Transactions on affective computing*, 4(2):127–141, 2013. 2, 7, 8

[9] Y. Li, S. Wang, Y. Zhao, and Q. Ji. Simultaneous facial feature tracking and facial expression recognition. *Image Processing, IEEE Transactions on*, 22(7):2559–2573, July 2013. 2, 3, 7

[10] P. Lucey, J. Cohn, T. Kanade, J. Saragih, Z. Ambadar, and I. Matthews. The extended cohn-kanade dataset (ck+): A complete dataset for action unit and emotion-specified expression. In *Computer Vision and Pattern Recognition Workshops (CVPRW), 2010 IEEE Computer Society Conference on*, pages 94–101, June 2010. 2, 6, 7

[11] I. Matthews and S. Baker. Active appearance models revisited. *International Journal of Computer Vision*, 60(2):135–164, Nov. 2004. 2

[12] G. McKeown, M. Valstar, R. Cowie, M. Pantic, and M. Schroder. The semaine database: Annotated multimodal records of emotionally colored conversations between a person and a limited agent. *IEEE Transactions on Affective Computing*, 3(1):5–17, 2012. 6, 7, 8

[13] S. Ren, X. Cao, Y. Wei, and J. Sun. Face alignment at 3000 fps via regressing local binary features. In *Computer Vision and Pattern Recognition (CVPR), 2014 IEEE Conference on*, pages 1685–1692, June 2014. 1, 2, 3

[14] T. Senechal, V. Rapp, H. Salam, R. Seguier, K. Bailly, and L. Prevost. Combining aam coefficients with lgbp histograms in the multi-kernel svm framework to detect facial action units. In *Automatic Face Gesture Recognition and Workshops (FG 2011), 2011 IEEE International Conference on*, pages 860–865, March 2011. 2, 8

[15] Y. Sun, X. Wang, and X. Tang. Deep convolutional network cascade for facial point detection. In *IEEE International Conference on Computer Vision and Pattern Recognition*, pages 3476–3483, 2013. 2

[16] M. Valstar, B. Jiang, M. Mehu, M. Pantic, and K. Scherer. The first facial expression recognition and analysis challenge. In *Automatic Face Gesture Recognition and Workshops (FG 2011), 2011 IEEE International Conference on*, pages 921–926, March 2011. 6, 7, 8

[17] M. Valstar and M. Pantic. Fully automatic recognition of the temporal phases of facial actions. *Systems, Man, and Cybernetics, Part B: Cybernetics, IEEE Transactions on*, 42(1):28–43, Feb 2012. 2, 7

[18] Z. Wang, Y. Li, S. Wang, and Q. Ji. Capturing global semantic relationships for facial action unit recognition. In *Computer Vision (ICCV), 2013 IEEE International Conference on*, pages 3304–3311, Dec 2013. 2, 7, 8

[19] X. Xiong and F. De la Torre Frade. Supervised descent method and its applications to face alignment. In *IEEE International Conference on Computer Vision and Pattern Recognition (CVPR)*, May 2013. 1, 2, 3, 6, 7, 8

[20] X. Zhang, M. Mahoor, S. Mavadati, and J. Cohn. A lp-norm mtmkl framework for simultaneous detection of multiple facial action units. In *Applications of Computer Vision (WACV), 2014 IEEE Winter Conference on*, pages 1104–1111, March 2014. 2, 7

[21] Z. Zhang, P. Luo, C. C. Loy, and X. Tang. Facial landmark detection by deep multi-task learning. In *The 13th European Conference on Computer Vision*, 2014. 3

[22] K. Zhao, W.-S. Chu, F. De la Torre, J. F. Cohn, and H. Zhang. Joint patch and multi-label learning for facial action unit detection. In *The IEEE Conference on Computer Vision and Pattern Recognition (CVPR)*, June 2015. 2